\documentclass[sigconf]{acmart}

\AtBeginDocument{%
  }

\renewcommand\footnotetextcopyrightpermission[1]{}
\setcopyright{acmcopyright}
\copyrightyear{2023}
\acmYear{2023}
\acmDOI{XXXXXXX.XXXXXXX}

\acmConference[MM' 23]{}{Oct 28-Nov 3,
  2023}{Ottawa, Canada}
\acmBooktitle{Proceedings of the 31st ACM International Conference on Multimedia (MM '23), October 28--November 3, 2023, Ottawa, Canada}
\acmPrice{15.00}
\acmISBN{978-1-4503-XXXX-X/18/06}

\acmSubmissionID{3019}

\usepackage{multirow}
\begin{document}

\title{Learning Dynamic Point Cloud Compression via Hierarchical Inter-frame Block Matching}

\author{Shuting Xia}
\authornote{Both authors contributed equally to this research.}
\email{xiashuting@sjtu.edu.cn}
\author{Tingyu Fan}
\authornotemark[1]
\email{woshiyizhishapaozi@sjtu.edu.cn}
\affiliation{%
  \department{Cooperative Medianet Innovation Center}
  \institution{Shanghai Jiaotong University}
  \city{Shanghai}
  \country{China}
}

\author{Yiling Xu}
\affiliation{%
  \department{Cooperative Medianet Innovation Center}
  \institution{Shanghai Jiaotong University}
  \city{Shanghai}
  \country{China}}
\email{yl.xu@sjtu.edu.cn}

\author{Jenq-Neng Hwang}
\affiliation{%
  \institution{University of Washington, Seattle}
  \country{USA}}
\email{hwang@uw.edu}

\author{Zhu Li}
\affiliation{%
 \institution{University of Missouri, Kansas City}
 \country{USA}}
\email{zhu.li@ieee.org}

\renewcommand{\shortauthors}{}

\begin{abstract}
3D dynamic point cloud (DPC) compression relies on mining its temporal context, which faces significant challenges due to DPC's sparsity and non-uniform structure. Existing methods are limited in capturing sufficient temporal dependencies. Therefore, this paper proposes a learning-based DPC compression framework via hierarchical block-matching-based inter-prediction module to compensate and compress the DPC geometry in latent space. Specifically, we propose a hierarchical motion estimation and motion compensation (Hie-ME/MC) framework for flexible inter-prediction, which dynamically selects the granularity of optical flow to encapsulate the motion information accurately. To improve the motion estimation efficiency of the proposed inter-prediction module, we further design a KNN-attention block matching (KABM) network that determines the impact of potential corresponding points based on the geometry and feature correlation. Finally, we compress the residual and the multi-scale optical flow with a fully-factorized deep entropy model. The experiment result on the MPEG-specified Owlii Dynamic Human Dynamic Point Cloud (Owlii) dataset shows that our framework outperforms the previous state-of-the-art methods and the MPEG standard V-PCC v18 in inter-frame low-delay mode.
\end{abstract}




\maketitle

\begin{figure*}[htbp]
  \centering
  \includegraphics[width=0.95\linewidth]{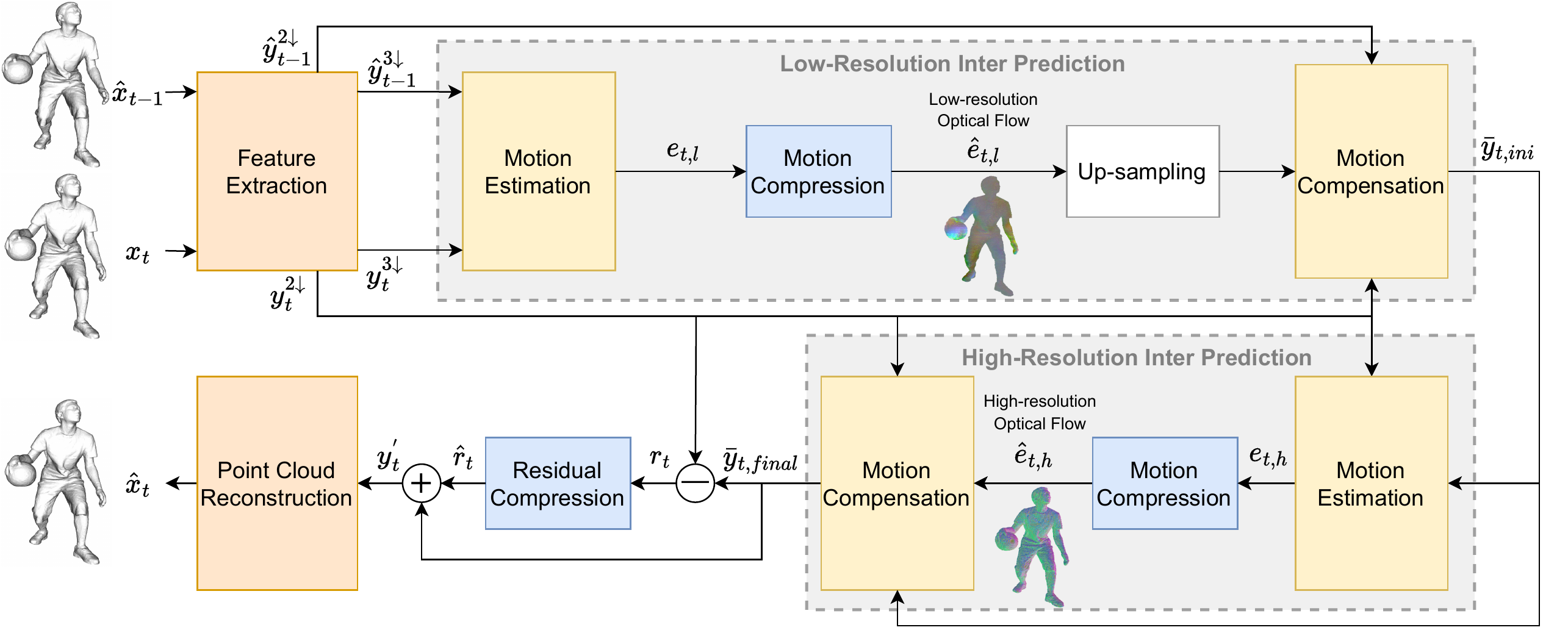}
  \caption{Overview of our proposed framework.}
  \label{figure-overview}  
\end{figure*}

\section{Introduction}
A dynamic point cloud (DPC) is a sequence of point cloud frames, where each frame is a set of unordered points sparsely distributed in the 3D space \cite{xu2018introduction}. DPC has become a promising 3D data format with broad applications in immersive media representation, volumetric content-capturing scenarios, and augmented and virtual reality (AR/VR) \cite{lizhu_summary}. However, DPC's huge data volume significantly burdens its storage and transmission, hindering its future development and application. Compared with the pixelized 2D image/video, DPC has a more flexible geometry structure, making its spatial and temporal dependencies more challenging to explore. Therefore, this paper focuses on DPC geometry compression, with efficient motion estimation and motion compensation (ME/MC) algorithms in feature space to capture the inter-frame dependencies.

Most existing DPC compression methods perform inter-prediction on consecutive frames based on hand-crafted temporal context selection algorithms. Some of these methods directly process DPC in 3D space by block-based matching, e.g., \cite{lz-FMV, lizhu-7, lizhu-10, block-based, Cluster-based} utilize different point cloud partitioning and matching algorithms to find point cloud regions' correspondence. Other DPC compression methods rely on projection algorithms \cite{helanyi, wenjie-view-dependent}. Among them, the Moving Picture Expert Group (MPEG) proposes Video-based Point Cloud Compression (V-PCC) \cite{emerging} that projects DPC into 2D geometry and texture videos, then leverages mature video codecs like HEVC to compress the generated videos. Among all the rule-based methods, V-PCC reports state-of-the-art performance.

Inspired by the application of deep-learning techniques in image and video compression, recent work begins to focus on neural point cloud compression. Instead of being limited to hand-crafted feature extraction and predictive mode selection, these methods are optimized throughout the whole dataset, thus achieving considerable gains against the rule-based algorithms. For static point cloud (SPC) compression, these learning-based methods can be concluded as point-based\cite{GRASP-Net, tianxin, linyao}, voxel-based\cite{jianqiang_lossy, jianqiang-multiscale, Andre}, and octree-based\cite{octsqueeze, octattention, VCN}, where the voxel-based method proposed by Wang et al.\cite{jianqiang-multiscale} achieves state-of-the-art performance on MPEG-specified 8iVFB\cite{8i} and Owlii\cite{owlii} datasets. Some explorations attempt to expand learning-based SPC compression to DPCs, e.g., Akhtar et al.\cite{anique} and Fan et al.\cite{tingyu-ddpcc}. However, the inter-prediction module in these methods either lacks explicit ME/MC structure \cite{anique} or relies on a simple block-matching mechanism and one-shot ME/MC \cite{tingyu-ddpcc}, which limits the motion estimation efficiency.

To address the above issues, we propose a learning-based DPC geometry compression network that adaptively matches blocks between the previously reconstructed and the current frame. Specifically, the network extracts hierarchical optical flow at different granularities, and analyses the local similarity of geometry and latent features to produce more accurate inter-frame matching. Our contributions are summarized as follows:

\begin{itemize}
\item We propose a hierarchical point cloud motion estimation and motion compensation (Hie-ME/MC) framework for inter-prediction, which supports adaptive granularity selection for optical flows of different magnitudes.
\item We design a KNN-attention block-matching (KABM) network to combine both geometry and feature correlation of probable corresponding points in our ME/MC module, which can strengthen the optical flow and improve the inter-prediction efficiency.
\item Our method achieves 88.80\% Bjontegaard Delta-Rate (BD-rate) gain against the MPEG standard V-PCC v18 in inter-frame low-delay mode and 9.96\% BD-rate gain against the previous state-of-the-art framework D-DPCC (Fan et al.\cite{tingyu-ddpcc}) on the Owlii\cite{owlii} dataset suggested by MPEG.
\end{itemize}

\section{Related Work}
\subsection{Rule-based DPC compression} 
Traditional DPC compression methods can be classified as 3D- and 2D-projection-based. 3D-based methods perform block matching to achieve ME/MC across various frames. \cite{lizhu-7} utilizes the octree structure to divide the point cloud iteratively and obtain the inter-frame correspondence via translation matrix estimation. \cite{lz-FMV} further introduce a half-voxel-level pattern to generate more accurate optical flows between octree blocks. Other work improves 3D motion estimation with different matching algorithms, e.g., Iterative closest points (ICP) \cite{block-based, lizhu-10, Cluster-based}, N-step Search \cite{block-based} and K-nearest-neighbor (KNN) Search \cite{Cluster-based}. On the other hand, 2D-projection-based methods leverage current video codecs, where most explorations focus on projection algorithm designing and patch arrangement. He et al. \cite{helanyi} utilize cubic projection. Zhu et al. \cite{wenjie-view-dependent} design a view-based hybrid global projection and patch-based local projection. Besides, MPEG proposes a powerful Video-based Point Cloud Compression (V-PCC) for DPC compression, which integrates cubic projection, patch generation, and packing for the geometry and texture map generation. Although the projection operation in 2D-projection-based methods inevitably introduces distortion that destroys inter-frame correspondence, V-PCC still achieves state-of-the-art performance among all 3D- and 2D-projection-based methods.

\subsection{Learning-based video compression}
Deep-learning-based video compression methods have recently emerged. DVC \cite{DVC} is the pioneering method of learning-based end-to-end video compression, which follows the residual-coding framework in H.264/AVC and H.265/HEVC but designs the modules using neural networks. Subsequent explorations refine DVC's inter-prediction module or entropy models, e.g., \cite{FVC} proposes feature-space motion compensation, and \cite{c2f} further introduces multi-scale motion estimation and quadtree-based block partition modes for entropy coding. In contrast, conditional-coding methods \cite{DCVC, DCVC-HEM, DCVC-DC} reduce temporal redundancies implicitly to better approximate the entropy lower bound. DCVC \cite{DCVC} first designs the conditional framework, which remains explicit ME/MC but generates temporal priors instead of  compensated feature maps to estimate the distribution parameters. DCVC-HEM \cite{DCVC-HEM} leverages a parallel-friendly dual-checkboard model for spatial prior. Moreover, DCVC-DC \cite{DCVC-DC} explores precise temporal context by group-wise offset estimation and multi-frame joint optimization, and allows a spatial-grouped entropy coding based on quadtree partition.

\subsection{Learning-based point cloud compression} 
Methods of learning-based point cloud compression have also emerged. For static point cloud compression, these methods can be roughly divided into point-based \cite{GRASP-Net, tianxin, linyao}, voxel-based\cite{jianqiang_lossy,jianqiang-multiscale, Andre} and octree-based\cite{octsqueeze, octattention, VCN}. Point-based methods utilize Farthest Point Sampling (FPS) and KNN to obtain local clusters and point-wise modules like PointNet++\cite{pointnet++} for spatial feature extraction. Voxel-based methods compress the voxelized point cloud as a 3D image, where Wang et al.\cite{jianqiang-multiscale} devise a multi-scale 3D sparseCNN-based network\cite{minkowski}, achieving remarkable gains against V-PCC (intra) on large-scale dense datasets like 8iVFB. Octree-based methods are octree entropy models that rely on spatial correlation exploration via ancestor or sibling nodes, which are targeted on extremely sparse point clouds like LiDAR point clouds.

For dynamic point cloud compression, Akhtar et al. \cite{anique} design a predictor that performs convolution on target coordinates to fuse the multi-scale features from the previous frame for motion compensation. However, Akhtar's network lacks an explicit ME/MC network to guide the inter-prediction. Therefore, Fan et al. \cite{tingyu-ddpcc} propose D-DPCC with an end-to-end feature-domain ME/MC module, which computes latent features of two successive frames for the subsequent motion estimation, and propose 3D adaptively weighted interpolation (3DAWI) for derivable motion compensation. Although D-DPCC outperforms other DPC compression frameworks, it performs inter-prediction only by one-shot ME/MC with a relatively simple motion estimation network, which remains with many temporal dependencies to explore.

\begin{figure}[!t]  
  \centering
  \includegraphics[width=0.93\linewidth]{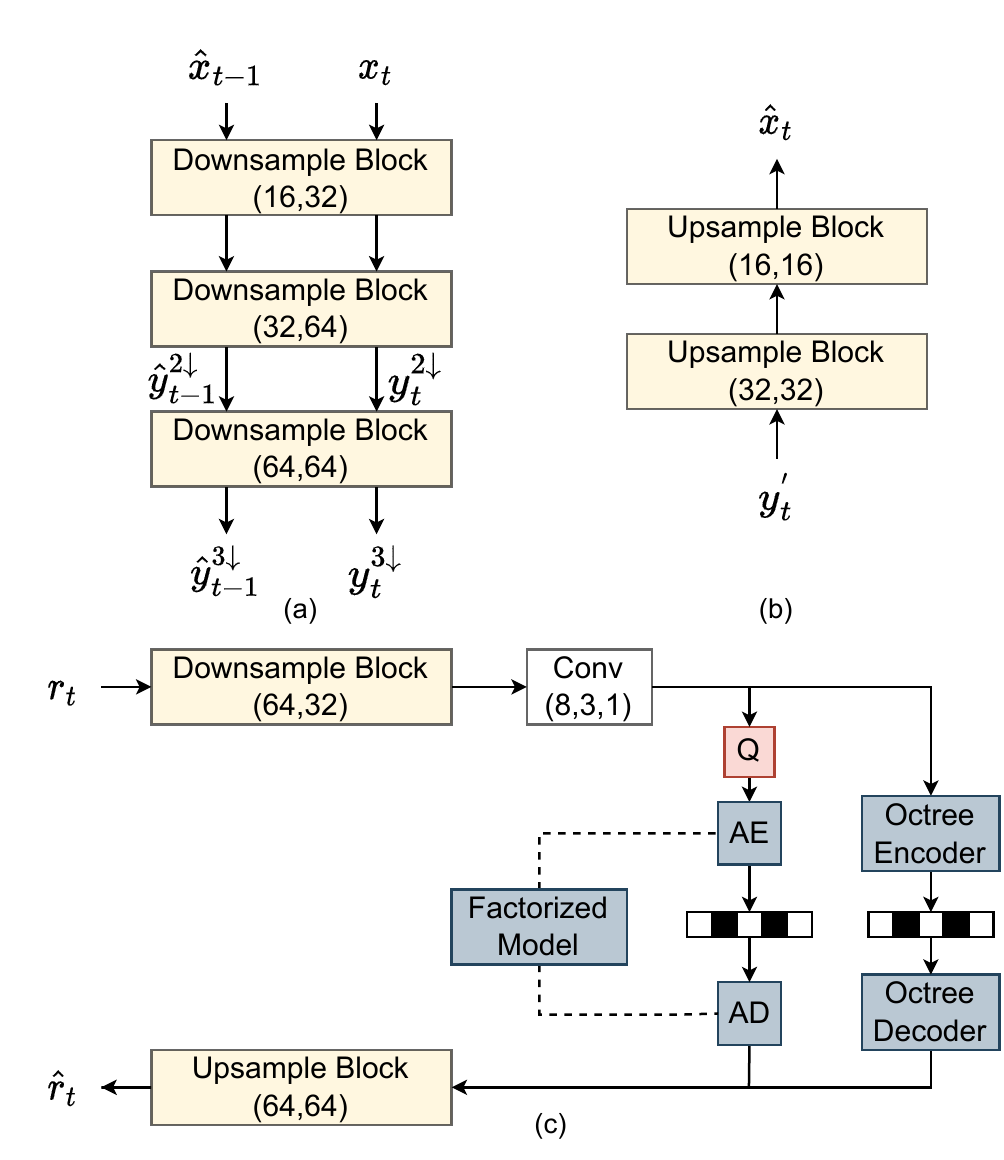}
  \caption{Frameworks of (a) Feature Extraction, (b) Point Cloud Reconstruction, (c) Residual Compression.}
  \label{feature-extraction}
\end{figure}

\section{Methods}
\subsection{Overview}
Figure \ref{figure-overview} shows that our network consists of five modules: feature extraction, low-resolution inter-prediction, high-resolution inter-prediction, residual compression, and point cloud reconstruction.  The input is processed as Minkowski sparse tensors \cite{minkowski} for complexity reduction like Wang et al. \cite{jianqiang-multiscale}.  Specifically, let $x_{t-1} = \left\{ C(x_{t-1}), F(x_{t-1}) \right\}$ and $x_t = \left\{ C(x_t), F(x_t) \right\}$ be two consecutive point cloud frames, where $ C(x_{t-1})$ and $C(x_t)$ are coordinates of occupied points, $ F(x_{t-1})$ and $F(x_t)$ are associated features with all-one values that indicate point occupancy. 

The network's input is the current frame $x_t$ and the previously reconstructed frame (reference frame) $\hat{x}_{t-1}$.  The feature extraction module encodes $x_t$ and $\hat{x}_{t-1}$ into latent features $y_t$ and $\hat{y}_{t-1}$ of different scales, where $y_t^{2 \downarrow}$/$\hat{y}_{t-1}^{2 \downarrow}$ and $y_t^{3 \downarrow}$/$\hat{y}_{t-1}^{3 \downarrow}$ separately indicate the $2\times$ and $3\times$ downsampled latent features. $y_t^{3 \downarrow}$ and $\hat{y}_{t-1}^{3 \downarrow}$ are the input of the low-resolution inter-prediction module, which generates the low-resolution flow embedding $\hat{e}_{t,l}$ and the initial prediction of $y_t^{2 \downarrow}$, i.e., $\bar{y}_{t, ini}$.  $\bar{y}_{t, ini}$ serves as the reference frame of the high-resolution inter-prediction that generates the high-resolution flow embedding $\hat{e}_{t,h}$ and the final prediction $\bar{y}_{t, final}$.  The residual compression module compresses and decompresses the residual $r_t$ between $y_t^{2 \downarrow}$ and $\bar{y}_{t, final}$. On the decoder side, $\bar{y}_{t, final}$ is added with the decompressed residual $\hat{r}_t$ to get the reconstructed latent feature $y_t'$.  $y_t'$ enters the point cloud reconstruction module to generate the reconstructed current frame $\hat{x}_t$.  In the following sections, we will introduce the architecture of each module in detail.

\begin{figure}[!t]  
  \centering
  \includegraphics[width=\linewidth]{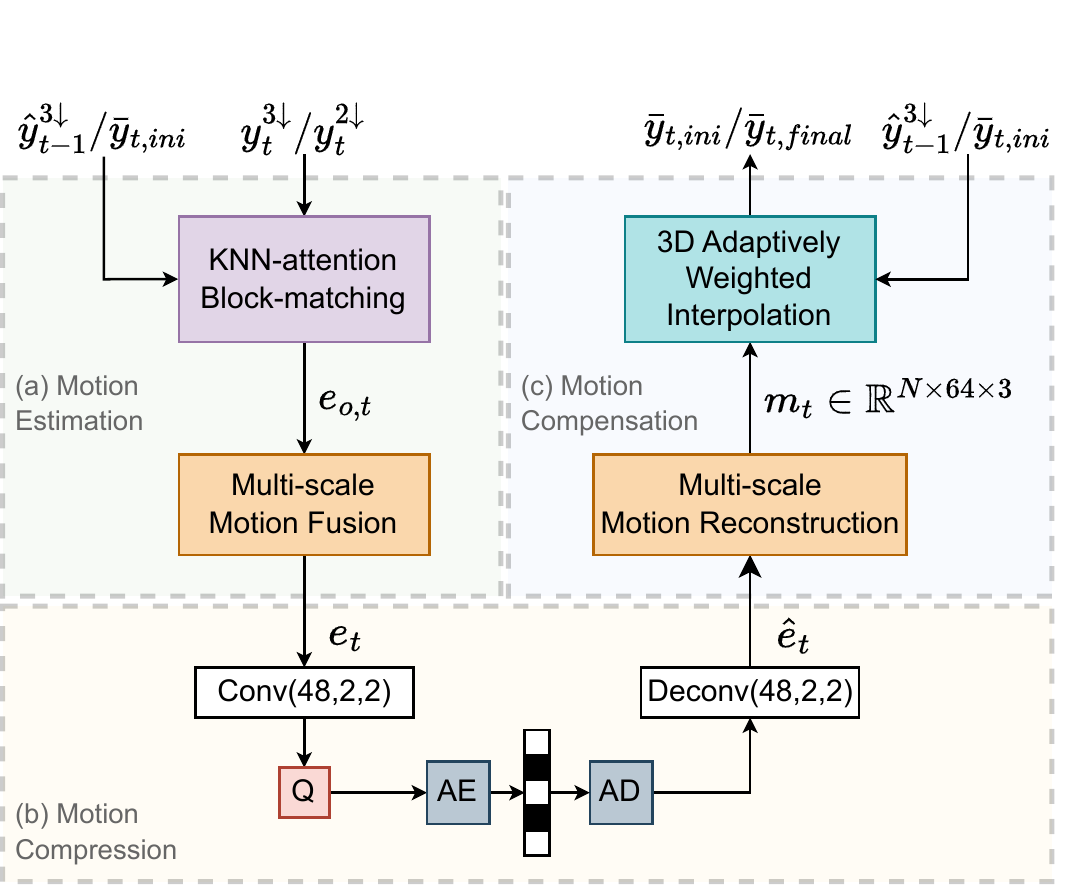}
  \caption{Frameworks of (a) Motion Estimation, (b) Motion Compression, (c) Motion Compensation.}
  \label{inter-prediction}
\end{figure}

\subsection{Feature Extraction}
As shown in Figure \ref{feature-extraction}(a), the feature extraction module follows \cite{tingyu-ddpcc} with serially connected down-sampling blocks \cite{jianqiang-multiscale} to hierarchically reduce spatial redundancies and produce latent features of $x_t$ and $\hat{x}_{t-1}$, i.e., $y_t$ and $\hat{y}_{t-1}$. Different from \cite{tingyu-ddpcc} with only $2\times$ down-sampling to generate single-scale latent features $y_t^{2 \downarrow}$/$\hat{y}_{t-1}^{2 \downarrow}$, this paper's feature extraction module contains one more down-sampling block to produce the $3\times$ down-sampled latent feature $y_t^{3 \downarrow}$/$\hat{y}_{t-1}^{3 \downarrow}$. The multi-scale latent features will be used for the subsequent hierarchical block-matching-based inter-prediction.

\begin{figure*}[htbp]
  \centering
  \includegraphics[width=0.95\linewidth]{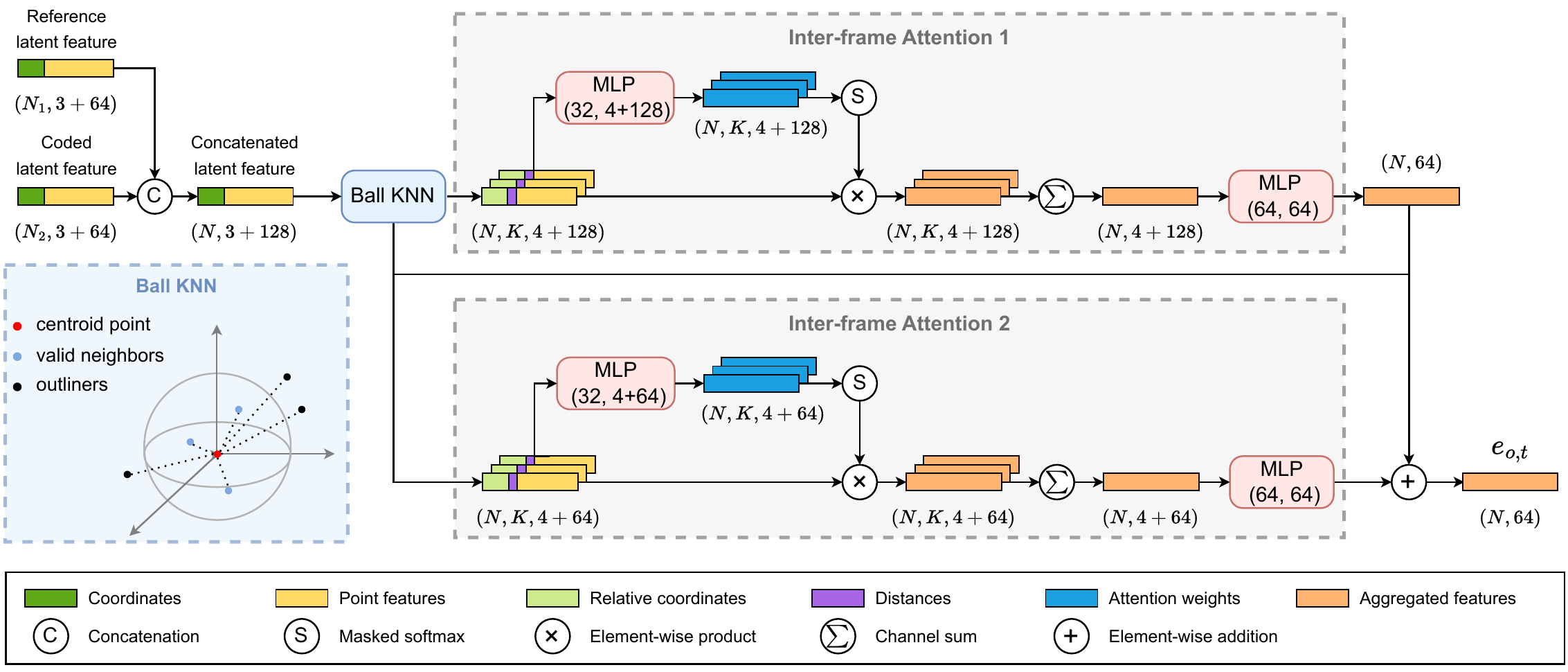}
  \caption{Feature space KNN-attention blocking-matching in motion estimation. MLP(H, O) means the cascade connection of a fully connected layer (outputs H channels), a ReLU function, and another fully connected layer (outputs O channels). }
  \label{point-based}  
\end{figure*}

\subsection{Hierarchical Block-matching-based Inter-prediction}
As shown in Figure \ref{inter-prediction}, we adopt the ME/MC framework in the one-shot inter-prediction module of D-DPCC \cite{tingyu-ddpcc}. Specifically, the motion estimation module analyses the temporal correlation between the latent feature $y_t$ and $\hat{y}_{t-1}$ with a KNN-attention block-matching (KABM) network and generates an original flow embedding $e_{o,t}$. $e_{o,t}$ passes through a Multi-scale Motion Fusion (MMF) \cite{tingyu-ddpcc} module to generate the multi-scale fused flow embedding $e_t$. The motion compression module is an Auto-Encoder-style network that compresses and decompresses $e_t$ with a non-parametric, fully factorized density model \cite{vae}, which will be introduced in \ref{residual compression} in detail. For motion compensation, the decompressed flow embedding $\hat{e}_t$ is fed into a Multi-scale Motion Reconstruction (MMR) module \cite{tingyu-ddpcc}. MMR restores and fuses the coarse- and fine-grained optical flow to generate the final optical flow $m_t$.

\noindent\textbf{Channel-wise Motion Compensation.}
The overall optical flow $m_t\in \mathbb{R}^{N\times 64\times 3}$ consists of 64 separate optical flows for every channel of the latent feature $y_t^{2 \downarrow} \in \mathbb{R}^{N\times 64}$, so that each channel can find its own match in the reference frame to provide richer temporal context. Specifically, the network first warps the coordinates in $y_t^{2 \downarrow}$ for each channel separately:
\begin{equation}\label{warp}
    u_w^{(i)} = u + m_{t,u}^{(i)}, \quad \forall u \in C(y_t^{2 \downarrow})
\end{equation}
where $C(y_t^{2 \downarrow})$ is the coordinates of $y_t^{2 \downarrow}$, $u$ denotes an arbitrary coordinate in $C(y_t^{2 \downarrow})$, $i$ is the channel index, $u_w^{(i)}$ is the warped coordinate of $u$ of the $i$-th channel, and $m_{t,u}^{(i)}=(\Delta x,\Delta y,\Delta z)$ is the optical flow located at $u$ of the $i$-th channel.

Subsequently, considering the sparse nature of point clouds and the derivability of prediction, we adopt the 3D Adaptively Weighted Interpolation (3DAWI) algorithm \cite{tingyu-ddpcc} for motion compensation:
\begin{equation}\label{3DAWI}
\bar{y}_{t,u}^{(i)} = \frac{ \sum_{v\in \vartheta(u_w^{(i)})} d(u_w^{(i)},v)^{-1} \cdot y_{ref,v}^{(i)} }{ \max{ \left( \sum_{v\in \vartheta(u_w^{(i)})} d(u_w^{(i)},v)^{-1},\alpha \right)} }, \quad \forall u \in C(y_t^{2 \downarrow})
\end{equation}
where $\vartheta(u_w^{(i)})$ is the 3-nearest-neighbor set of $u_w^{(i)}$, $d(u_w^{(i)},v)^{-1}$ is the inverse Euclidean distance between $u_w^{(i)}$ and the neighbor $v$, and $y_{ref,v}^{(i)}$ is the feature value of the reference frame $y_{ref}$ defined at position $v$ and channel $i$. In the low-resolution inter-prediction, $y_{ref}$ is $\hat{y}_{t-1}^{2 \downarrow}$, and in the high-resolution inter-prediction, $y_{ref}$ is $\bar{y}_{t, ini}$. $\alpha$ is a hyperparameter named \textit{penalty coefficient} that adaptively decreases the weight of isolated warped points. Note that $C(y_t^{2 \downarrow})$ is required in Equation \ref{warp} and \ref{3DAWI}, which is losslessly coded (See more details in the Appendices).

\noindent\textbf{Hierarchical ME/MC.}
Inspired by the rate-distortion optimization (RDO) algorithms in MPEG video compression standards \cite{VVC}, which adaptively selects the optimal block size for motion estimation, we transfer the idea of two-stage ME/MC from the deep video codec proposed by Lu et al. \cite{c2f} to our DPC codec, which enables more flexible choices of the granularity of the optical flow. 

Figure \ref{figure-overview} illustrates the structure of the hierarchical ME/MC module (Hie-ME/MC). The framework is divided into two parts. First, the low-resolution inter-prediction takes $3\times$ down-sampled latent features of the previous and current frame, i.e., $\hat{y}_{t-1}^{3 \downarrow}$ and $y_t^{3 \downarrow}$ as input to generate the low-resolution optical flow $e_{t,l}$, which describes coarse and large movements between blocks. In other words, the lower-scale motion estimation captures the "base" movements of a large block of points, corresponding to larger macroblock or coding unit partitions in AVC, HEVC, and VVC. The reconstructed low-resolution optical flow $\hat{e}_{t,l}$ is up-sampled to match the scale of $\hat{y}_{t-1}^{2 \downarrow}$ and $y_t^{2 \downarrow}$ for motion compensation. Finally, the network performs a rough compensation from $\hat{y}_{t-1}^{2 \downarrow}$ to $y_t^{2 \downarrow}$  with the low-resolution flow embedding $\hat{e}_{t,l}$ and produces an initial prediction $\bar{y}_{t, ini}$ for $y_t^{2 \downarrow}$. 

The high-resolution inter-prediction inherits the architecture of low-resolution. However, $\bar{y}_{t, ini}$ instead of $\hat{y}_{t-1}^{2 \downarrow}$ serves as the reference frame to estimate the high-resolution flow embedding $e_{t,h}$. Since the low-resolution ME/MC has already captured coarse movements at the low resolution ($3\times$ downsampled), this stage further extracts finer-grained motion at a higher resolution ($2\times$ downsampled) and outputs the final predicted latent representation $\bar{y}_{t, final}$ for $y_t^{2 \downarrow}$.

\noindent\textbf{KNN-attention Block-matching.}
The previous work D-DPCC about DPC compression \cite{tingyu-ddpcc} designs a simple 2-layer SparseCNN for motion estimation, which fails to calculate the explicit correspondence of blocks between two frames. Therefore, we design a KNN-attention block-matching (KABM) network for more accurate motion estimation. As Figure \ref{point-based} depicts, the KABM consists of one ball-KNN search and two inter-frame attention networks. At first, the network concatenates the reference frame and the current frame, so that blocks in both frames can aggregate information simultaneously. The concatenation operation of sparse tensors is defined as follows:
\begin{equation}\label{concate}
y_{cat, u} = 
\begin{cases}
y_{1, u} \oplus y_{2, u} & \mbox{u}\in C(y_1)\cap C(y_2) \\
y_{1, u} \oplus \textbf{0} & \mbox{u}\in C(y_1), \text{u}\notin C(y_2) \\
\textbf{0} \oplus y_{2, u} & \mbox{u}\notin C(y_1), \text{u}\in C(y_2),
\end{cases}
\end{equation}
where $y_1$ and $y_2$ are latent features of the reference and current frames, $C(y_1)$ and $C(y_2)$ are corresponding integer coordinates, and $\oplus$ means the concatenation of feature channels. Then, the network uses ball-KNN to find the neighbors of $y_{cat}$ limited in a spherical region (the maximum number of neighbors is K) and generates the neighbors' attributes, including relative coordinates, distances, and features. After that, the network uses two inter-frame attention networks to aggregate features between two frames and output an original flow embedding $e_{o,t}$ that encapsulates the matching information. Each inter-frame attention network generates an attention weight matrix for the neighbors based on their attribute vectors and uses softmax to normalize the weight matrix. Note that instead of the "hard" match in FlowNet3D \cite{flownet3d} that preserves only the most correlated point, we use the weight matrix to achieve "soft" matching because the motion vector can point to the intersection of two or more blocks. Finally, the weighted sum of neighbors' attributes passes through another MLP to achieve channel-wise aggregation, producing the final output $e_{o,t}$.

\subsection{Residual Compression} \label{residual compression}
As Figure \ref{feature-extraction}(c) shows, the residual Compression module encodes the residual $r_t$ between $y_t^{2 \downarrow}$ and the final prediction $\bar{y}_{t, final}$ with an Auto-Encoder (AE) style network. On the encoder side, a downsample block \cite{jianqiang-multiscale} and a convolution layer serve as the parametric analysis transform that transforms $r_t$ into a more compact latent representation $l_{r_t}=\left\{ C(y_t^{3 \downarrow}), F(l_{r_t}) \right\}$. The coordinate $C(y_t^{3 \downarrow})$ is losslessly compressed by G-PCC octree v14 \cite{emerging}, and the features $F(l_{r_t})$ is quantized and coded using a non-parametric, fully factorized density model \cite{vae}. On the decoder side, an upsample block serves as the parametric synthesis transform to recover the reconstructed residual $\hat{r}_t$. The compression of fused flow embedding $e_t$ follows a similar process, as Figure \ref{inter-prediction}(b) shows, where a convolution/transpose convolution layer serves as the encoder/decoder.

\subsection{Point Cloud Reconstruction} \label{point cloud reconstruction}
The point cloud reconstruction module contains upsample blocks \cite{jianqiang-multiscale,tingyu-ddpcc} symmetric to the feature extraction, which recovers the current frame hierarchically $\hat{x}_t$ from the reconstructed latent feature $y'_t$, as depicted in Figure \ref{feature-extraction}(b).  In addition, a sparse convolution layer with one output channel is utilized to generate the occupancy possibilities of each point. We adopt an adaptive pruning strategy \cite{jianqiang-multiscale} to remove false points based on the occupancy possibility.

\begin{figure}[!t]  
  \centering
  \includegraphics[width=0.75\linewidth]{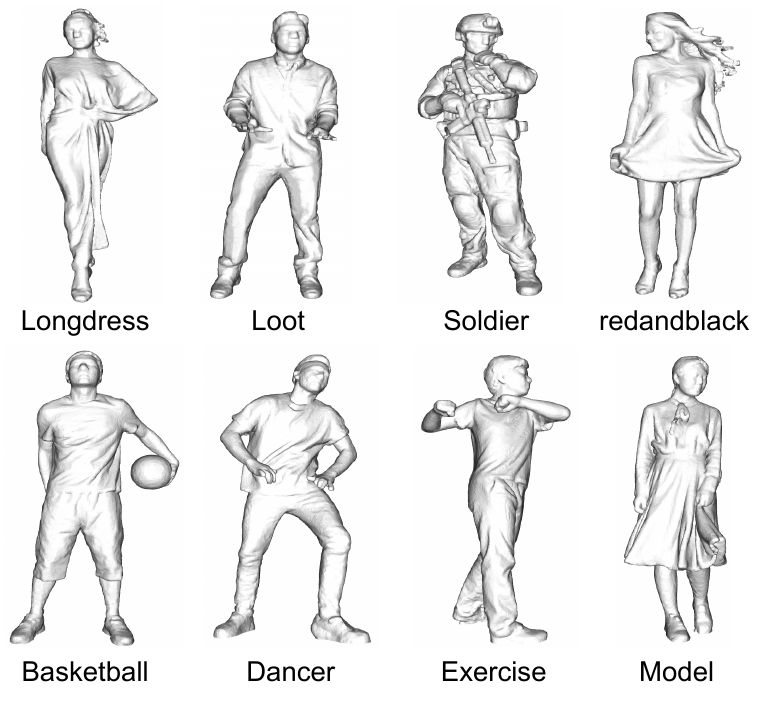}
  \caption{Visualization of 8iVFB (\emph{Longdress}, \emph{loot}, \emph{soldier} and \emph{redandblack}) and Owlii dataset (\emph{Basketball}, \emph{Dancer}, \emph{Exercise} and \emph{Model}).}
  \label{dataset}
\end{figure}

\begin{figure*}[htbp]  
  \centering
  \includegraphics[width=\linewidth]{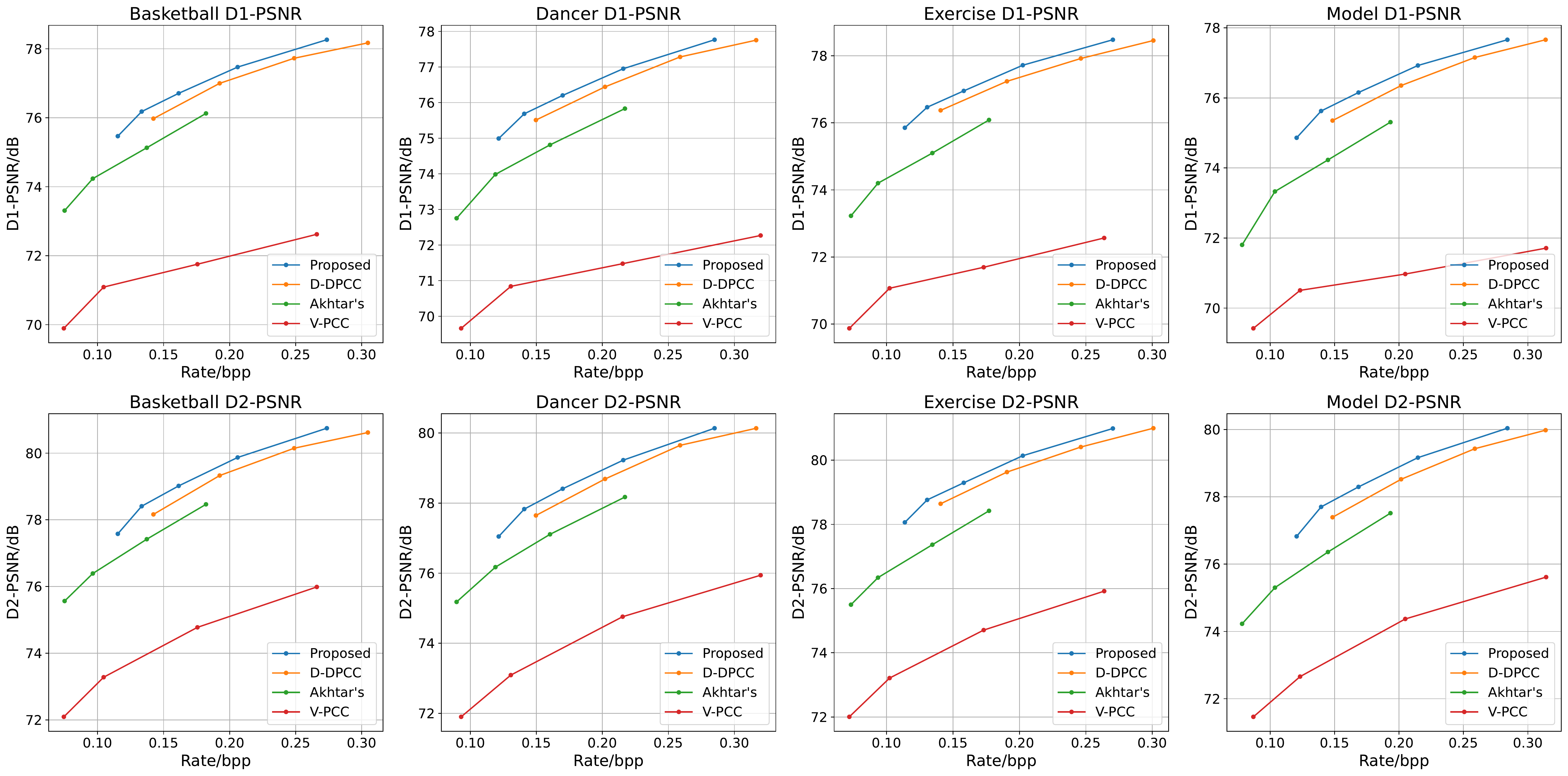}
  \caption{Rate-Distortion curves on Owlii (Basketball, Dancer, Exercise, Model) test sequences with D1 (point-to-point) and D2 (point-to-plane) PSNR distortion measurements. }
  \label{results}
\end{figure*}

\subsection{Loss Function}
We utilize the Lagrangian rate-distortion loss function to optimize our network end-to-end, as shown in Equation \ref{loss} :
\begin{equation}\label{loss}
  \mathcal{L} = \mathcal{R}_l+\mathcal{R}_h+\mathcal{R}_r + \lambda \mathcal{D},
\end{equation}
where $\mathcal{R}_l$ and $\mathcal{R}_h$ are the bits per point (bpp) for the coding of the low-resolution flow embedding $\hat{e}_{t,l}$ and the high-resolution flow embedding $\hat{e}_{t,h}$, $\mathcal{R}_r$ represents the bpp of encoding the residual $\hat{r}_t$, $\mathcal{D}$ is the distortion between $x_t$ and the reconstructed current frame $\hat{x}_t$, and $\lambda$ is a parameter to control the trade-off between bit rate and distortion.

\noindent\textbf{Rate.}
Let $F$ be one latent feature generated by the analysis transform. First, $F$ needs to be quantized before encoding. During training, the quantization is approximated by adding a uniform noise $w \sim \mathcal{U}(-0.5, 0.5)$ to ensure differentiability. Second, the quantized feature $\tilde{F}$ is encoded using arithmetic coding with a fully factorized entropy model \cite{vae}, which estimates the probability distribution of $\tilde{F}$, i.e., $P_{\tilde{F} \textbar \psi}$, where $\psi$ are learnable parameters in the factorized model. The bpp of $F$ is calculated as:
\begin{equation}
    \mathcal{R}_{\tilde{F}} = \frac{1}{N} \sum_i -\log_{2}{\left( P_{\tilde{F}_i \textbar \psi^{(i)}} \right)},
\end{equation}
where $N$ is the number of points in the original point cloud $x_t$, and $i$ is the channel index. 

\noindent\textbf{Distortion.}
As mentioned in \ref{point cloud reconstruction}, the proposed upsample blocks estimate the occupancy possibility $p_v$ of each point $v$ with a sparse convolution layer. We utilize the Binary Cross Entropy (BCE) to measure the distortion:
\begin{equation}
    \mathcal{D}_{BCE} = \frac{1}{N} \sum_v -\left( \mathcal{O}_v \log{p_v} + (1-\mathcal{O}_v) \log{(1-p_v)} \right),
\end{equation}
where $\mathcal{O}_v$ is the ground truth that whether the point $v$ is truly occupied (1) or not (0) in $x_t$. In hierarchical reconstruction, BCEs of upsample blocks are averaged as the final distortion $D$:
\begin{equation}
    \mathcal{D} = \frac{1}{K} \sum_{k=1}^{K} \mathcal{D}_{BCE}^{k}, 
\end{equation}
where $k$ is the scale index.

\section{EXPERIMENTS}
\subsection{Experiment Settings}
\noindent\textbf{Datasets.} 
We train the proposed network following the MPEG AI-3DGC EE5.3 common test condition (CTC) using the 8i Voxelized Full Bodies (8iVFB) dataset \cite{8i}, which contains four sequences with 1200 frames. Each sequence's frame rate is 30 frames per second (fps) over 10 seconds, and the spatial precision is 10 bits. For evaluation, we use the Owlii Dynamic Human DPC dataset \cite{owlii} as CTC suggests, containing four sequences with 2400 frames, whose frame rate is 30 fps over 20 seconds for each sequence. As CTC requires, the first 100 frames of each sequence in Owlii are quantized from 11 bits to 10 bits and tested in our experiment.

\noindent\textbf{Training Details.} 
We train five models for different bit rates with $\lambda = 5, 6, 7, 10, 15$. We use an Adam optimizer with $\beta = (0.9, 0.999)$ and a linear learning-rate scheduler in which the learning rate decays at the rate of 0.7 for every 15 epochs. Each model is trained for 100 epochs with a two-stage training strategy: for the first ten epochs, $\lambda$ is set as 20 to accelerate the convergence of the network; then $\lambda$ recovers its predefined value for the remaining epochs. The batch size is set as 4. All experiments are conducted on a GeForce RTX 3090 GPU with 24GB memory. For 3DAWI interpolation, the \textit{penalty coeffecient} $\alpha$ is set as 3. For ball-KNN, the radius $r=3$ and $K=16$.

\begin{table}[!t]
\begin{center}
\caption{BD-rate(\%) gains against D-DPCC, Akhtar et al. and V-PCC(inter). Negative values of BD-rate represent the percentage of bit rate reduction.}
\label{result_table}
\begin{tabular}{l l|c c c c c c}
\toprule[1pt]
\multicolumn{2}{l|}{\multirow{2}{*}{Sequence}} & \multicolumn{2}{c}{D-DPCC} & \multicolumn{2}{c}{Akhtar's} & \multicolumn{2}{c}{V-PCC}\\
\multicolumn{2}{l|}{} & D1 & D2 & D1 & D2 & D1 & D2\\
\midrule
\multicolumn{2}{l|}{Basketball} & -9.56 & -10.01 & -27.55 & -24.05 & -83.84 & -66.34\\
\multicolumn{2}{l|}{Dancer} & -9.04 & -8.33 & -31.61 & -27.02 & -74.38 & -70.22\\
\multicolumn{2}{l|}{Exercise} & -10.05 & -9.38 & -33.15 & -31.48 & -97.13 & -69.20\\
\multicolumn{2}{l|}{Model} & -11.18 & -11.30 & -32.72 & -29.48 & -99.85 & -70.79\\
\midrule
\multicolumn{2}{l|}{\textbf{Average}} & \textbf{-9.96} & \textbf{-9.75} & \textbf{-31.26} & \textbf{-28.00} & \textbf{-88.80} & \textbf{-69.12}\\
\bottomrule[1pt]
\end{tabular}
\end{center}
\end{table}

\begin{figure}[!t]  
  \centering
  \includegraphics[width=\linewidth]{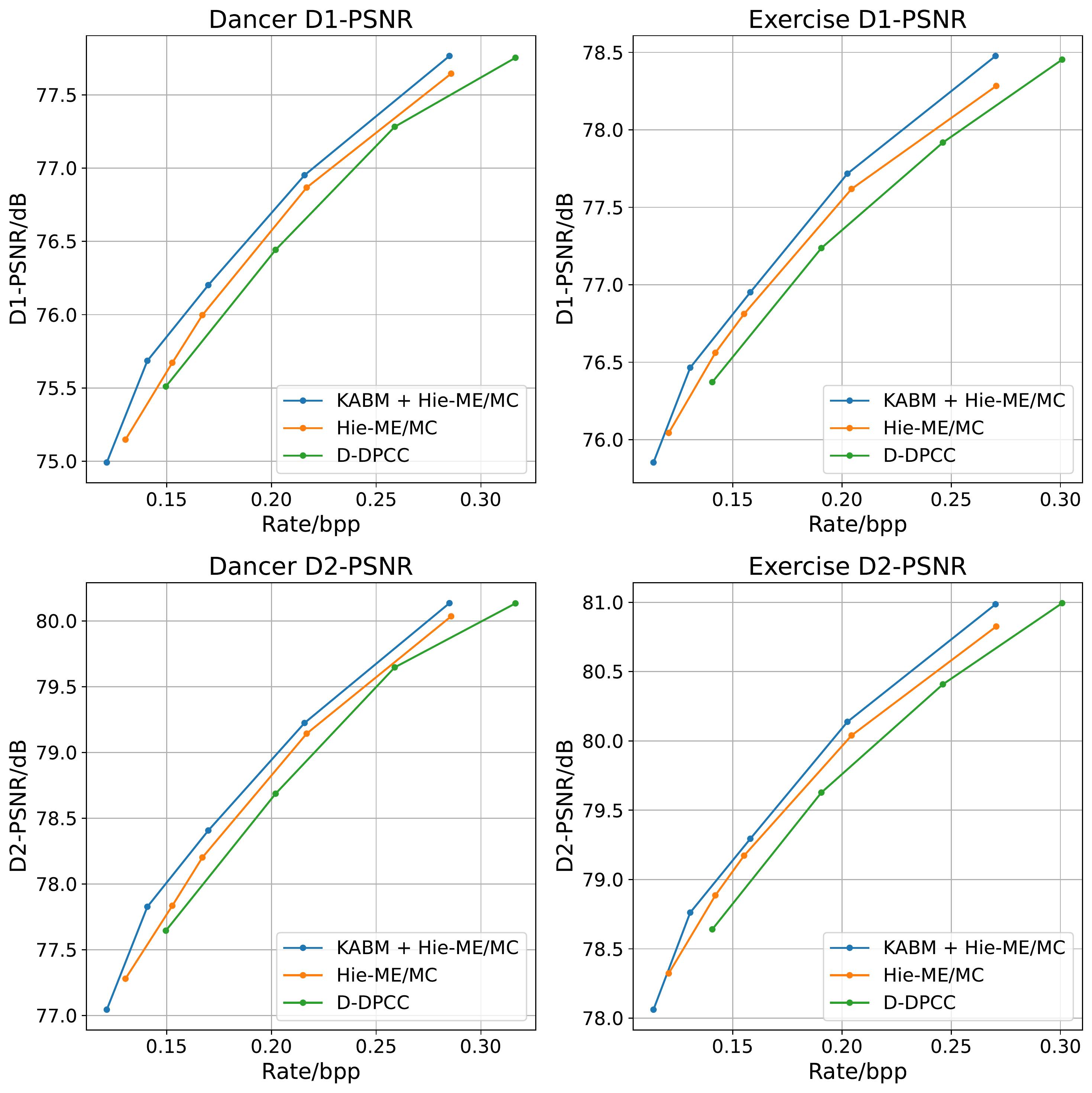}
  \caption{Effect of Hie-ME/MC and KABM.}
  \label{ablation}
\end{figure}

\noindent\textbf{Evalution Metrics.} 
Following the objective measurements in MPEG CTC, the bit rate is evaluated by bits per point (bpp), and the distortion is measured by point-to-point (D1) and point-to-plane (D2) distance-based Peak Signal-to-Value (PSNR). The peak value is 1023 for quantized Owlii. We plot rate-distortion curves and calculate the Bjontegaard Delta-Rate (BD-rate) gains of our proposed method against previous state-of-the-art methods, representing the average bit rate reduction percentage at the same objective quality. The first frame (I-frame) of each sequence is intra-coded by PCGCv2 \cite{jianqiang-multiscale}, and the following frames (P-frame) are coded based on the previously reconstructed frame.

\begin{table}[!t]
\begin{center}
\caption{Ablation studies about Hie-ME/MC and KABM.}
\label{ablation_table}
\begin{tabular}{l l|c c c c}
\toprule[1pt]
\multicolumn{2}{l|}{\multirow{3}{*}{Sequence}} & \multicolumn{2}{c}{Hie-ME/MC} & \multicolumn{2}{c}{KABM+Hie-ME/MC}\\
\multicolumn{2}{l|}{} & \multicolumn{2}{c}{vs D-DPCC} & \multicolumn{2}{c}{vs Hie-ME/MC}\\
\multicolumn{2}{l|}{} & D1 & D2 & D1 & D2\\
\midrule
\multicolumn{2}{l|}{Dancer} & -4.87 & -4.66 & -4.56 & -3.96\\
\multicolumn{2}{l|}{Exercise} & -5.92 & -5.93 & -4.12 & -3.43\\
\bottomrule[1pt]
\end{tabular}
\end{center}
\end{table}

\noindent\textbf{Baseline Settings.} 
We compare our proposed method with the previous DPC geometry compression methods D-DPCC \cite{tingyu-ddpcc} and Akhtar's framework \cite{anique}, as well as the MPEG standard V-PCC test model v18 (inter-frame low-delay mode). For a fair comparison, we re-train D-DPCC with the dataset mentioned above. For Akhtar's framework, we obtain their official result from the related MPEG proposal \cite{anique_data} because they also adopt the MPEG CTC. For V-PCC, we select the geometry QP in video coding as 24, 20, 16, and 12.

\subsection{Experimental results}
Figure \ref{results} plots the rate-distortion curves on Owlii test sequences with D1-PSNR and D2-PSNR objective quality metrics, and the corresponding BD-rate comparison is shown in Table \ref{result_table}. First, the learning-based methods all outperform the state-of-the-art traditional DPC codec V-PCC due to the optimization over the whole dataset, where our method achieves 88.80\% (D1) and 69.12\% (D2) average BD-rate gains on four sequences. Compared with Akhtar's framework, our method has an explicit hierarchical ME/MC architecture with optical flows to guide the inter-prediction, leading to 31.26\% (D1) and 28.00\% (D2) average BD-rate gains. Compared with the previous state-of-the-art method D-DPCC, our method achieves 9.96\% (D1) and 9.75\% (D2) average BD-rate gains due to the proposed two-stage ME/MC and the KNN-attention block-matching module, which allow more accurate inter-prediction and leverages richer temporal context. It is worth mentioning that our method has even higher BD-rate gains on sequences with smaller movements, i.e., \emph{Exercise} and \emph{Model}, as Table \ref{result_table} illustrates, demonstrating our model's significant ability to capture fine-grained movements due to the proposed Hie-ME/ME and KABM.

\subsection{Analysis} \label{analysis}
\noindent\textbf{Ablation Study.}
Figure \ref{ablation} shows the ablation studies on sequence \emph{dancer} and \emph{exercise} that demonstrates the effectiveness of Hie-ME/MC and KABM. The Hie-ME/MC brings a 5\% BD-rate gain compared with D-DPCC (one-stage ME/MC) due to its hierarchical ME/MC architecture. Besides, the network with both KABM and Hie-ME/MC achieves 4\% BD-rate gain compared with the network with only Hie-ME/MC. We attribute this to the attention weights in KABM that explicitly measures the correspondence of neighbors. Note that all ablation studies are done with the same settings except for the inter-prediction module. Therefore, the ablation studies demonstrate that both Hie-ME/MC and KABM can increase inter-prediction efficiency, leading to an increase in overall performance. 

\begin{figure}[!t]  
  \centering
  \includegraphics[width=\linewidth]{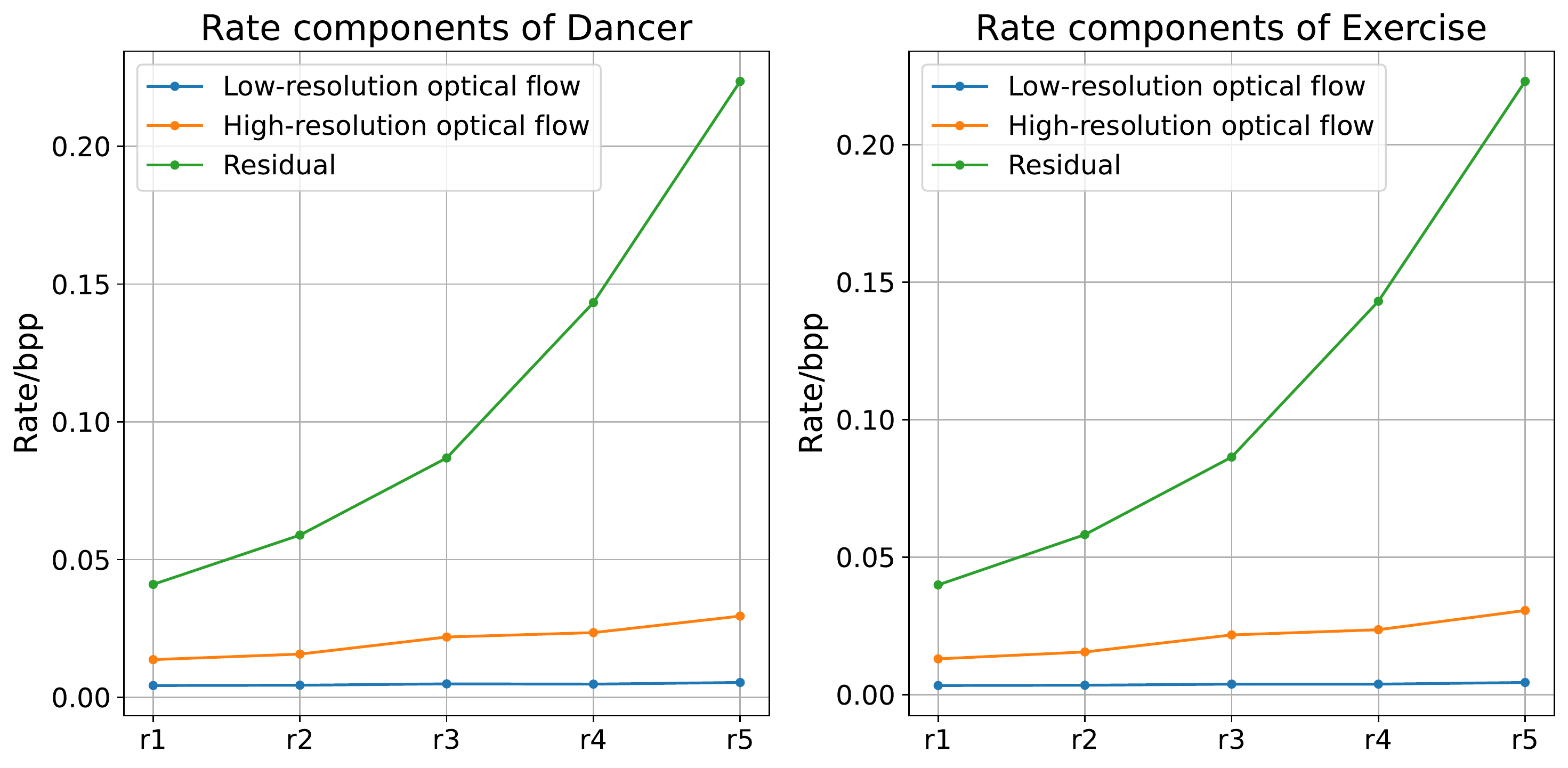}
  \caption{Rate components of different bit rates. $r1$, $r2$, $r3$, $r4$ and $r5$ represent five different total bit rates corresponding to models trained with $\lambda = 5, 6, 7, 10, 15$,}
  \label{rate}
\end{figure}

\begin{figure*}[htbp]  
  \centering
  \includegraphics[width=0.95\linewidth]{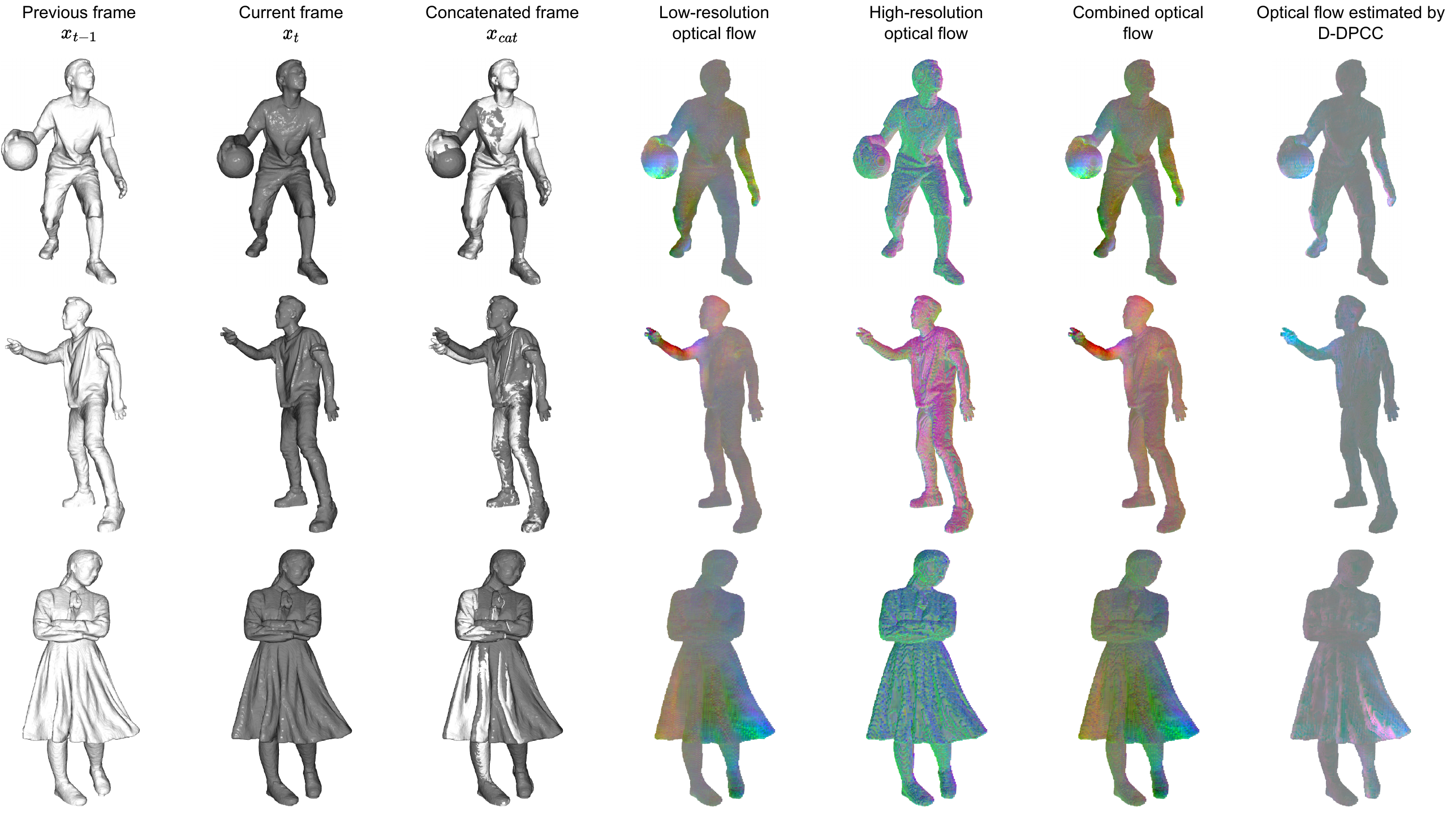}
  \caption{Visualization of the multi-resolution optical flows estimated by our framework and the single-resolution optical flows estimated by D-DPCC. The optical flow $(\Delta x, \Delta y, \Delta z)$ is normalized and re-scaled to (0, 255) to serve as RGB values in the visualization, where gray represents no optical flow, and other colors represent optical flows of different directions and magnitudes. Rows from top to bottom: Sequence \emph{Basketball}, \emph{Dancer}, and \emph{Model}.}
  \label{visual}
\end{figure*}

\noindent\textbf{Motion bit rate consumption.}
Figure \ref{rate} shows the analysis of the bit rate components. The total rate $\mathcal{R}$ has three components: bits for encoding the residual, low-resolution, and high-resolution optical flows, i.e., $\mathcal{R}_r$, $\mathcal{R}_l$ and $\mathcal{R}_h$. We use $r1-r5$ to represent the five different bit rate levels from low to high. In general, $\mathcal{R}$ and $\mathcal{R}_r$ increases as the parameter $\lambda$ increases. Besides, as the overall bit rate increases, the two frames' latent features $y_t$ and $\hat{y}_{t-1}$ encapsulate more fine-grained information, which slightly increases $\mathcal{R}_h$ because it captures the "finetuning" motion information. In comparison, $\mathcal{R}_l$ is almost unchanged because it captures the "base" movements between large blocks, which remains almost constant regardless of the fineness of $y_t$ and $\hat{y}_{t-1}$.

\noindent\textbf{Optical Flow Visualization.} To demonstrate the effectiveness of Hie-ME/MC, we visualize the low-resolution optical flow, high-resolution optical flow, the combination of the two-resolution flows, and the optical flow estimated by one-stage ME/MC in D-DPCC. Note that the inter-prediction module learns to output all the optical flow end-to-end without any ground-truth data about optical flow during training.
We can observe that low-resolution optical flow mainly captures large movements, e.g., hitting the basketball and waving hands. With the low-resolution inter-prediction capturing "base" movements, high-resolution optical flow mainly focuses on the movements of texture details, like the deformation of clothing. Besides, compared with the optical flow estimated by D-DPCC with one-shot ME/MC, the Hie-ME/MC and KABM encourage the optical flow estimation. We can observe that the combined optical flow estimated by our network is richer and more detailed than the optical flow estimated by D-DPCC.

\subsection{Model Complexity}
As shown in Table \ref{time}, we calculate and compare the runtime of the methods mentioned above. The runtime is evaluated using Intel(R) Xeon(R) Gold 6226R CPU @2.90GHz and a GeForce RTX 3090 GPU with 24GB memory.
We can observe that V-PCC v18 has the highest complexity because V-PCC needs to perform operations like patch segmentation, patch packing, and image generation, leading to excessive complexity and a lack of parallelism. Akhtar's method is the fastest due to the lack of an explicit ME/MC structure, which also leads to worse performance than methods with ME/MC modules, as shown in Figure \ref{results}. For the two DPC compression frameworks, the runtime of our framework is slightly longer compared with D-DPCC due to the introduction of one more inter-prediction stage in the Hie-ME/MC and KABM module. However, the overall increment of complexity is insignificant because inter-prediction only takes up a small percentage of time in the whole encoding and decoding process.

\begin{table}[!t]
\begin{center}
\caption{Runtime of different methods on Owlii.}
\label{time}
\begin{tabular}{l|c c c c}
\toprule[1pt]
{Time(s/frame)} & {Proposed} & {D-DPCC} & {Akhtar's} & {V-PCC}\\
\midrule
{Enc} & {0.87} & {0.67} & {0.47} & {81.01}\\
{Dec} & {0.87} & {0.72} & {0.68} & {1.25}\\
\bottomrule[1pt]
\end{tabular}
\end{center}
\end{table}

\section{Conclusion}
This paper proposes a learning-based dynamic point cloud geometry compression framework via hierarchical inter-frame block matching. We introduce a two-stage hierarchical motion estimation and motion compensation (Hie-ME/MC) module for inter-prediction, which adaptively extracts multi-resolution optical flows to reduce temporal redundancies. We also design a KNN-attention block-matching (KABM) module for aggregating local geometry and feature correlations, achieving flexible and precise motion estimation. Our proposed method achieves average 9.96\% BD-rate gain against the previous state-of-the-art method D-DPCC and 88.80\% BD-rate gain against the MPEG standard V-PCC Test Model v18 on Owlii dataset following the MPEG Common Test Conditions.


\bibliographystyle{ACM-Reference-Format}
\bibliography{ms}  

\clearpage 

\appendix

\section{Network Architecture}
This section provides detailed network architectures of several modules in our proposed framework.

\subsection{Downsample Block and Upsample Block}
In our framework, the Downsample Blocks and Upsample Blocks constitute the Feature Extraction module and Point Cloud Reconstruction module like \cite{tingyu-ddpcc}, which generate multiscale latent features and hierarchically recover the coded frame respectively. They are also utilized for the Auto-Encoder style Residual Compression module, as shown in Figure \ref{feature-extraction}. 

\begin{figure}[!t]  
  \centering
  \includegraphics[width=\linewidth]{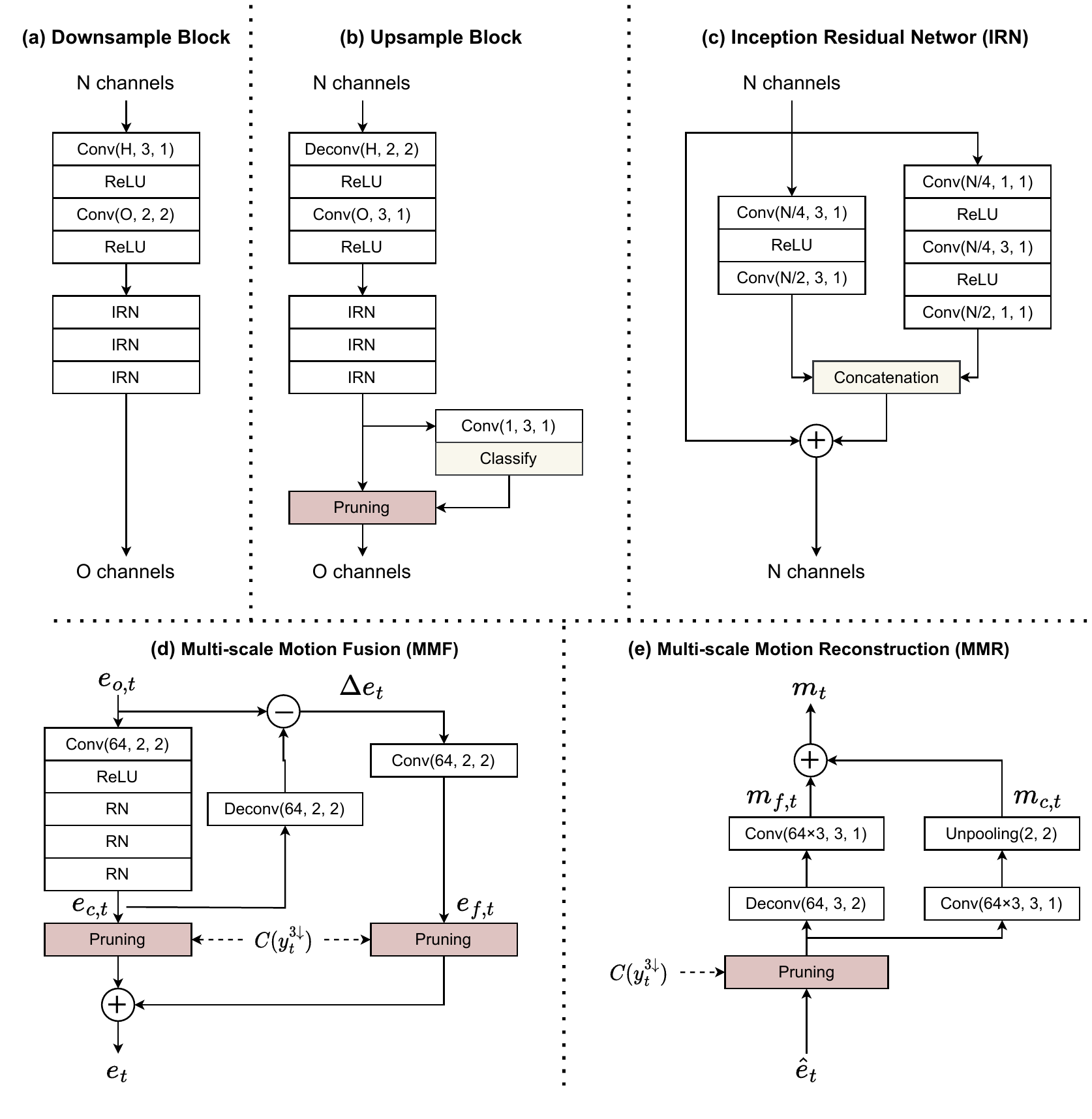}
  \caption{The architecture of (a) Downsample Block (H, O), (b) Upsample Block (H, O), (c) Inception Residual Network (IRN), (d) Multi-scale Motion Fusion (MMF), (e) Multi-scale Motion Reconstruction (MMR).
  H and O indicate the number of hidden and output channels.}
  \label{temp}
\end{figure}

The detailed structure of Downsample Block and Upsample Block is shown in Figure \ref{temp}(a)(b). Specifically, Downsample Block consists of a stride-two sparse convolution layer for point cloud downsampling, as well as three serially connected Inception Residual Network (IRN) blocks (Figure \ref{temp}(c)) for local feature extraction and aggregation. Upsample Block mirrors Downsample Block with a stride-two sparse transpose convolution layer for point cloud upsampling, followed by three IRNs. Moreover, Upsample Block has classification and pruning operations to remove false points based on the estimated occupancy possibilities generated by an output-channel-one sparse convolution layer. Points with lower occupancy possibility are regarded as needless and pruned.

\subsection{Multi-scale Motion Fusion and Reconstruction}
In Hie-ME/MC, we adopt the ME/MC framework in the one-stage inter-prediction module of D-DPCC \cite{tingyu-ddpcc} as shown in Figure \ref{inter-prediction}, in which a Multi-scale Motion Fusion (MMF) module is utilized to generate a multiscale fused flow embedding $e_t$ in Motion Estimation and a Multi-scale Motion Reconstruction (MMR) module is used to generate the final optical flow $m_t$ in Motion Compensation. The architecture of MMF and MMR is shown in Figure \ref{temp}(d)(e).

MMF analyzes the temporal correlation between the previous reconstructed frame $\hat{x}_{t-1}$ and current frame $x_t$ by further downsampling the original flow embedding $e_{o,t}$ generated by KABM. However, more downsampling operations lead to more information loss. To solve this problem, MMF remains both coarse- and fine-grained flow information. It first downsamples $e_{o,t}$ by a stride-two sparse convolution layer to enlarge the perception field, followed by three Residual Networks (RNs) to produce a coarse-grained flow embedding $e_{c,t}$. Then the residual between $e_{c,t}$ and $e_{o,t}$, i.e., $\Delta e_t$, is calculated and downsampled to generate a fine-grained flow embedding $e_{f,t}$, which compensates for the information loss during the downsampling from $e_{o,t}$ to $e_{c,t}$. Finally, $e_{c,t}$ and $e_{f,t}$ are pruned and added to generate the multi-scale fused flow embedding $e_t$.

MMR is symmetrical to MMF, which recovers the coarse- and fine-grained optical flow, i.e., $m_{c,t}$ and $m_{f,t}$, by an unpooling and a stride-two sparse transpose convolution layer respectively. And $m_{c,t}$ and $m_{f,t}$ are also summed up to generate the final optical flow $m_t$. Note that $m_{c,t}$, $m_{f,t}$ and $m_t \in \mathbb{R}^{N\times 64\times 3}$, including 64 separate optical flows for each channel of the latent feature $y_t^{2 \downarrow} \in \mathbb{R}^{N\times 64}$, so that each channel can find its own corresponding neighbors in the reference frame, get richer temporal contexts and more precise prediction.

\begin{figure*}[htbp]  
  \centering
  \includegraphics[width=\linewidth]{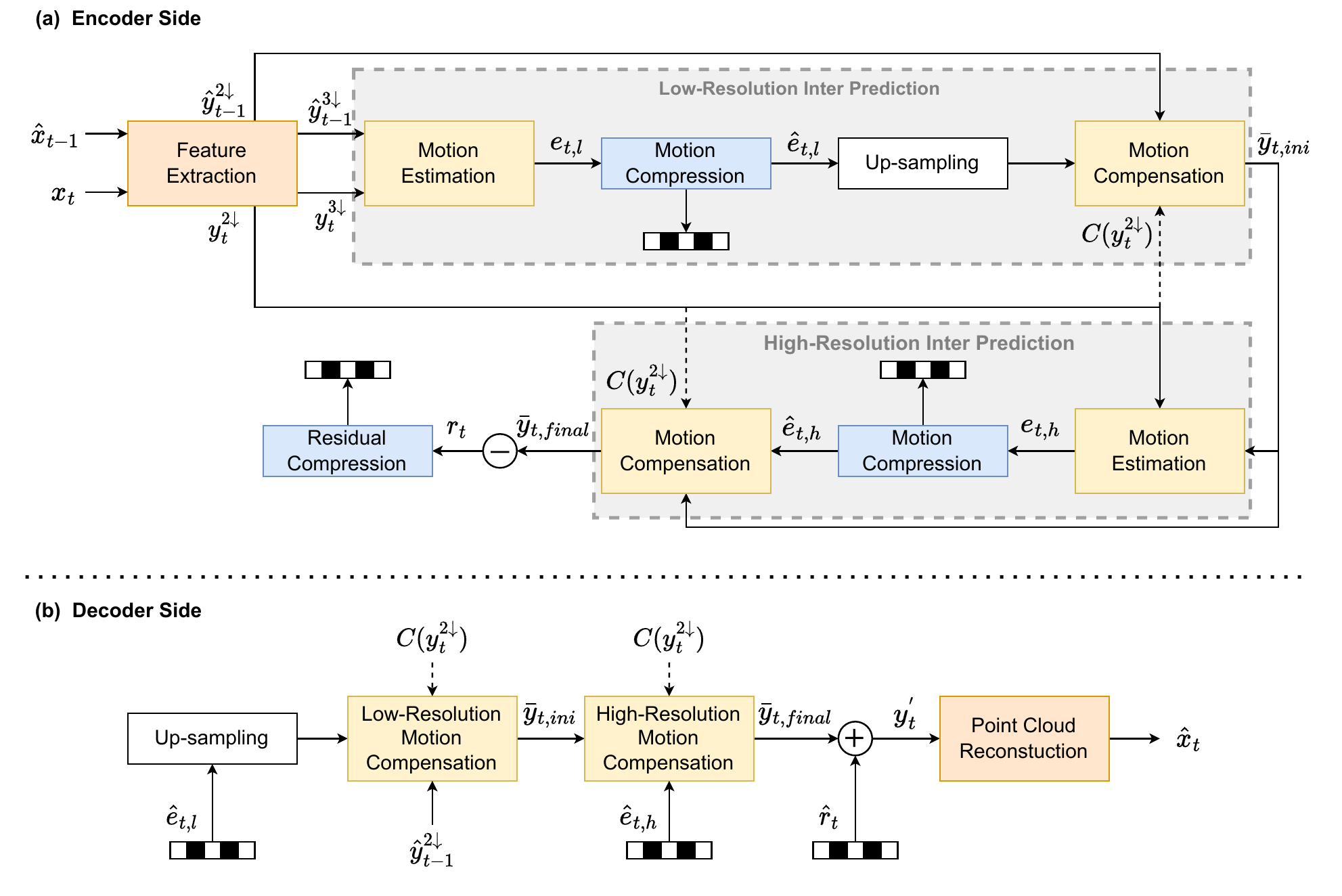}
  \caption{Practical encoding and decoding process.}
  \label{encdec}
\end{figure*}

\section{Practical Encoding and Decoding}
This section introduces the practical operations needed for encoding and decoding. We will show what information is written into bitstreams and how the network reconstructs the current frame.

\subsection{On the Encoder Side and Decoder Side}
Figure \ref{encdec} illustrates the practical process of encoding and decoding. In particular, on the encoder side, the Feature Extraction module generates $2\times$ and $3\times$ down-sampled latent features of the previous reconstructed frame $\hat{x}_{t-1}$ and current frame $x_t$, denoted as $\hat{y}_{t-1}^{2\downarrow}$/$y_t^{2\downarrow}$ and $\hat{y}_{t-1}^{3\downarrow}$/$y_t^{3\downarrow}$. The $3\times$ down-sampled latent features enter the low-resolution inter-prediction module to generate low-resolution optical flow $e_{t,l}$, which is compressed (as Figure \ref{inter-prediction}(b)) and written into the bitstream. The encoder side then decompresses the bitstream and obtains reconstructed low-resolution optical flow $\hat{e}_{t,l}$. $\hat{e}_{t,l}$ is up-sampled, compensating for the base movements between $\hat{y}_{t-1}^{2\downarrow}$ and $y_t^{2\downarrow}$. Motion compensation only need coordinates of $y_t^{2\downarrow}$ instead of the whole $y_t^{2\downarrow}$, and generates an initial prediction of $y_t^{2\downarrow}$, i.e., $\bar{y}_{t, ini}$. High-resolution inter-prediction follows similar procedures to low-resolution, while the movements between $\bar{y}_{t, ini}$ and $y_t^{2\downarrow}$ are estimated, and the final prediction $\bar{y}_{t, final}$ is produced. High-resolution optical flow $\hat{e}_{t,h}$ is also compressed and transmitted to the decoder side. Finally, the encoder side compress and transmit the residual (denoted as $r_t$) between $\bar{y}_{t, final}$ and $y_t^{2\downarrow}$.

For the decoder side, it only utilizes Motion Compensation modules in low- and high-resolution inter-prediction. The decoder side first entropy decodes and reconstructs the low- and high-resolution optical flows, i.e., $\hat{e}_{t,l}$ and $\hat{e}_{t,h}$. Given $C(y_t^{2\downarrow})$ and $\hat{e}_{t,l}$, the warped coordinates of points in $y_t^{2\downarrow}$ can be calculated and the initial prediction of $y_t^{2\downarrow}$ is generated by 3DAWI, as Eq. \ref{warp} and \ref{3DAWI} show. The generation of the final prediction $\bar{y}_{t, final}$ follows the same operations. After that, the reconstructed residual $\hat{r}_t$ is summed up with $\bar{y}_{t, final}$ to produce the reconstructed latent feature $y'_t$. In the end, the Point Cloud Reconstruction module up-samples $y'_t$ to generate the reconstruction $\hat{x}_t$ of the current frame.

For a dynamic point cloud sequence with several frames, the first frame (I-frame) is intra-coded by PCGCv2 \cite{jianqiang-multiscale}, and the subsequent frames (P-frames) are coded by our framework based on the previous reconstructed frame.

\begin{figure}[!t]  
  \centering
  \includegraphics[width=\linewidth]{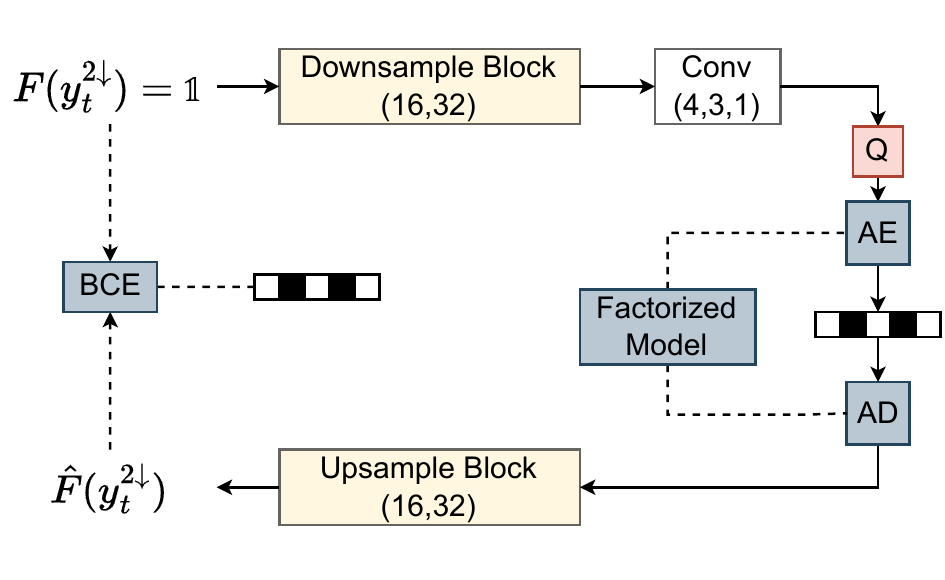}
  \caption{Lossless compressor for $C(y_t^{2\downarrow})$. $F(y_t^{2\downarrow})\in \mathbb{R}^{N\times 1}$ describes the occupancy of each point in $C(y_t^{2\downarrow})$, which is an all-one vector.}
  \label{lossless}
\end{figure}

\subsection{Lossless Compressor for $C(y_t^{2 \downarrow})$}
As described above, the decoder side needs the coordinates of the $2\times$ down-sampled latent feature of the current frame in Motion Compensation, i.e., $C(y_t^{2\downarrow})$. As shown in Figure \ref{lossless}, we losslessly code $C(y_t^{2\downarrow})$ with an Auto-Encoder-style network similar to the Residual Compression module (Figure \ref{feature-extraction}(c)). Note that the $3\times$ down-sampled coordinates $C(y_t^{3\downarrow})$ is losslessly coded in the Residual Compression module using G-PCC octree v14 \cite{emerging}. So we only need to compress the occupancy of $2\times$ down-sampled coordinates and can obtain $C(y_t^{2\downarrow})$ by up-sampling $C(y_t^{3\downarrow})$ and pruning based on occupancy. Specifically, we input an all-one vector representing the occupancy of points in $C(y_t^{2\downarrow})$, denoted as $F(y_t^{2\downarrow})=\textbf{1}, \textbf{1} \in \mathbb{R}^{N\times 1}$. It passes through a parametric analysis transform (a Downsample Block and a sparse convolution layer) to generate a more compact latent representation, which is quantized and coded by the arithmetic codec with a fully factorized density model \cite{vae}. Then an Upsample Block serves as the parametric synthesis transform that recovers a lossy reconstruction of $F(y_t^{2\downarrow})$, i.e., $\hat{F}(y_t^{2\downarrow})$. The Binary Cross Entropy (BCE) between $F(y_t^{2\downarrow})$ and $\hat{F}(y_t^{2\downarrow})$ is also calculated for lossless compression, which describes the bitrate of coding $F(y_t^{2\downarrow})$ losslessly with the distribution estimated by the fully factorized density model.


\end{document}